\documentclass{article}
\usepackage{spconf,graphicx,amssymb}
\usepackage{amsmath}
\usepackage{booktabs} 
\usepackage{multirow}
\usepackage{multicol}
\usepackage[T1]{fontenc}
\usepackage{algorithm}
\usepackage{algpseudocode}
\usepackage{color, soul}
\usepackage{hyperref}

\usepackage[table,xcdraw]{xcolor}
\usepackage{setspace}
\usepackage{enumitem}
\DeclareUnicodeCharacter{2212}{-}

\algnewcommand{\LeftComment}[1]{\Statex \(\triangleright\) #1}

\title{Sem-CS: Semantic CLIPStyler for Text-Based Image Style Transfer}
\name{Chanda Grover Kamra$^{\star}$ \qquad Indra Deep Mastan$^{\dagger}$ \qquad Debayan Gupta$^{\star}$}  
  \address{$^{\star}$ Ashoka University, India. $^{\dagger}$ The LNM Institute of Information Technology (LNMIIT), India.
      }

\begin{document}

\maketitle
\begin{abstract}
CLIPStyler demonstrated image style transfer with realistic textures using only a style text description (instead of requiring a reference style image). However, the ground semantics of objects in the style transfer output is lost due to style spill-over on salient and background objects (content mismatch) or over-stylization. To solve this, we propose Semantic CLIPStyler (Sem-CS), that performs semantic style transfer.

Sem-CS first segments the content image into salient and non-salient objects and then transfers artistic style based on a given style text description. The semantic style transfer is achieved using global foreground loss (for salient objects) and global background loss (for non-salient objects).
Our empirical results, including DISTS, NIMA and user study scores, show that our proposed framework yields superior qualitative and quantitative performance. Our code is available at \href{https://github.com/chandagrover/sem-cs}{github.com/chandagrover/sem-cs}.


\end{abstract}

%

\section{Introduction}
\label{sec:intro}
Image style transfer~\cite{gatys2016image, NEURIPS2021_df535469, li2017universal, park2019arbitrary, mastan2022dilie, mechrez2018contextual, 8451734} aims to synthesize new images by transferring style features such as colour and texture patterns to the content image. Image style transfer can be classified into photo-realistic style transfer~\cite{BMVC2017_153, Luan2017DeepPS} and artistic style transfer~\cite{gatys2016image, samuth2022patch} based on the input content image and style image. One problem in image style transfer is the fact that a user needs to find a good reference image with the desired style.

\begin{figure}[!ht]
    \centering
    \begin{minipage}{0.19\linewidth}
     \centering
        \textbf{Style Text}
    \end{minipage}
    \hfill
    \begin{minipage}{0.19\linewidth}
     \centering
        \textbf{Input Image}
    \end{minipage}
    \hfill
    \begin{minipage}{0.19\linewidth}
     \centering
        CLIP-Styler~\cite{Kwon_2022_CVPR}
    \end{minipage}
    \hfill
    \begin{minipage}{0.19\linewidth}
     \centering
        Gen-Art~\cite{yang2022generative}
    \end{minipage}
    \hfill
    \begin{minipage}{0.19\linewidth}
     \centering
        Sem-CS (ours)
    \end{minipage}
    \fbox{
    \begin{minipage}[c][1.3cm][c]{1.3cm}
     \centering
     \small
      A Monet style painting
    \end{minipage}}
        \begin{minipage}{0.19\linewidth}
         \centering
             \includegraphics[width=0.99\linewidth]{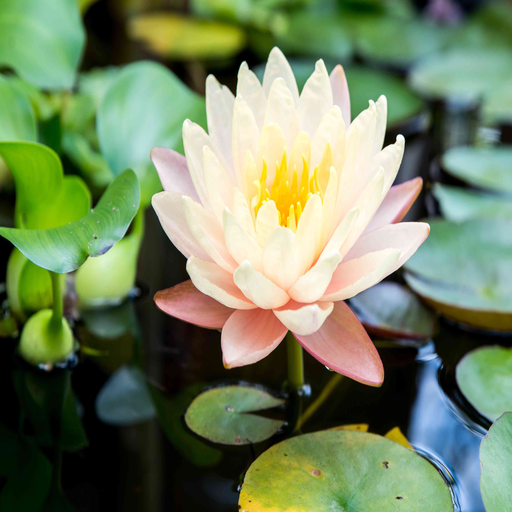}
        \end{minipage}
        \hfill
        \begin{minipage}{0.19\linewidth}
         \centering
             \includegraphics[width=0.99\linewidth]{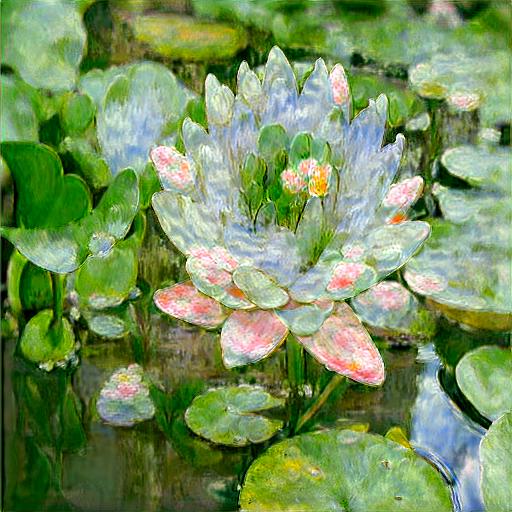}
        \end{minipage}
        \begin{minipage}{0.19\linewidth}
         \centering                 
            \includegraphics[width=0.99\linewidth]{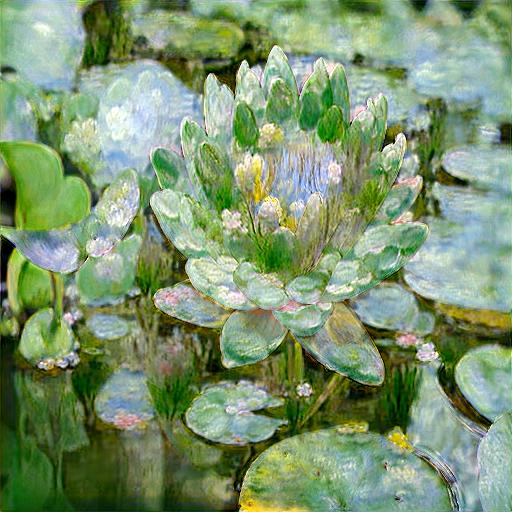}
        \end{minipage}
        \begin{minipage}{0.19\linewidth}
         \centering                 
            \includegraphics[width=0.99\linewidth]{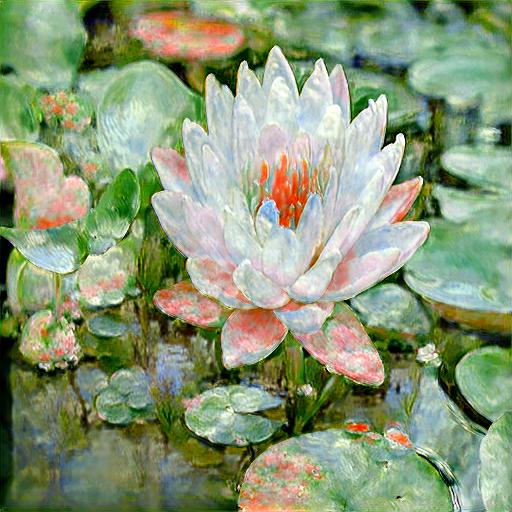}
        \end{minipage}
   \vspace{0.2cm}     
   \hrule
   \vspace{0.2cm}
    \fbox{
   \begin{minipage}[c][1.3cm][c]{1.3cm}
     \centering
     \small
      Desert Sand
    \end{minipage}}
    \hfill
    \begin{minipage}{0.19\linewidth}
     \centering
      \includegraphics[width=0.99\linewidth]{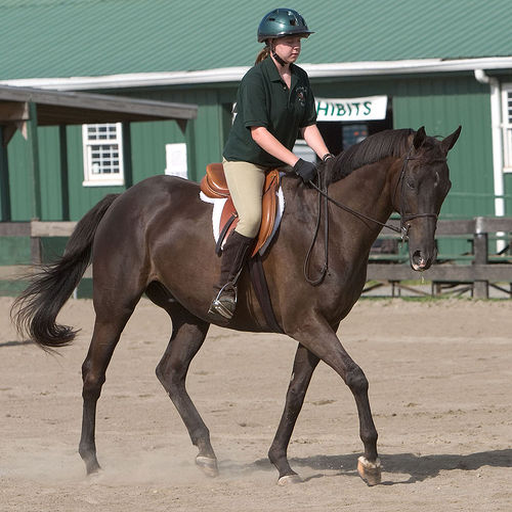}
    \end{minipage}
    \begin{minipage}{0.19\linewidth}
     \centering
      \includegraphics[width=0.99\linewidth]{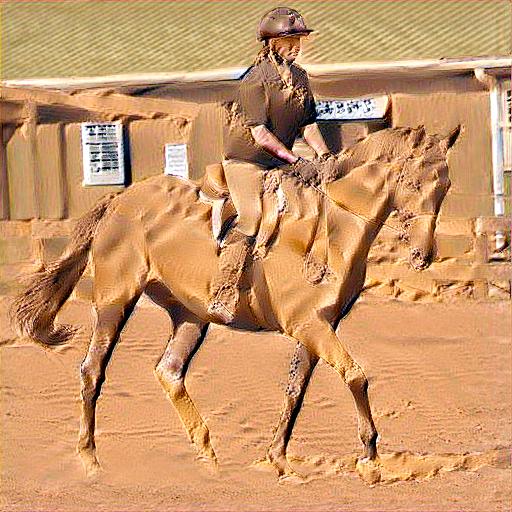}
    \end{minipage}
    \hfill
    \begin{minipage}{0.19\linewidth}
     \centering
      \includegraphics[width=0.99\linewidth]{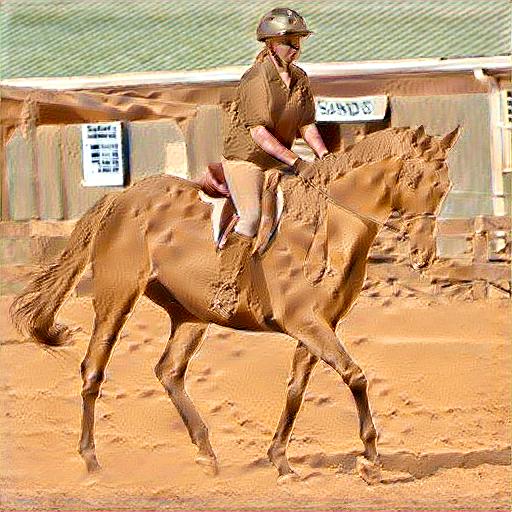}
    \end{minipage}
    \begin{minipage}{0.19\linewidth}
     \centering
      \includegraphics[width=0.99\linewidth]{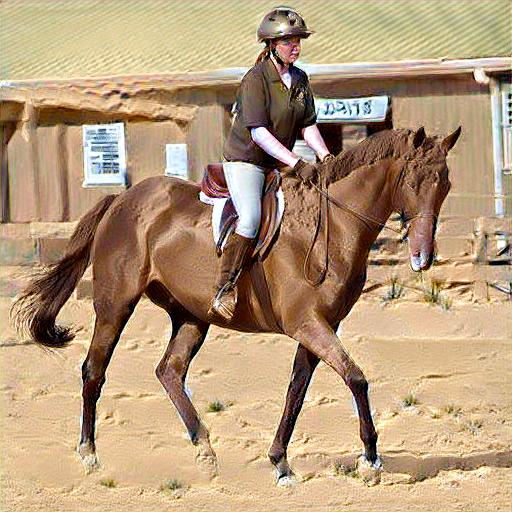}
    \end{minipage}
    \caption{\small{The figure illustrates over-stylization and the effects of content mismatch on style transfer output. \textbf{Top row} CLIPStyler~\cite{Kwon_2022_CVPR} and Gen-Art~\cite{yang2022generative} over-stylize style features on salient objects and image background as the content features of the flower are lost. Sem-CS (ours) preserved the semantics of the flower. \textbf{Bottom row} CLIPStyler~\cite{Kwon_2022_CVPR} and Generative Artisan~\cite{yang2022generative} outputs suffer from content mismatch as the Desert Sand style is applied to both man and horse. Sem-CS (ours) performed style transfer while minimizing content mismatch and preserving semantics.}}
    \label{fig:main_fig}
\end{figure}

Recently, CLIPStyler~\cite{Kwon_2022_CVPR} proposed a novel artistic style transfer approach that uses a text condition to perform style transfer without a reference style image. However, it suffers from the over-styling problem, which results in the distortion of content features in the output image (Fig.~\ref{fig:main_fig}-first row).


Another challenge in style transfer is when style spillover between dissimilar objects occurs, also known as the content mismatch problem~\cite{Luan2017DeepPS} (Fig.~\ref{fig:main_fig}-first-second row). Content mismatch reduces the visual quality of the style transfer output, and it is hard to avoid when the semantic objects in the style and the content features are of different types and numbers~\cite{Mastan_2021_CVPR, context2022icvgip}. A good style transfer approach minimizes both content mismatch and over-styling.

    \begin{figure*}[!h]
        \includegraphics[width=1.05\textwidth]{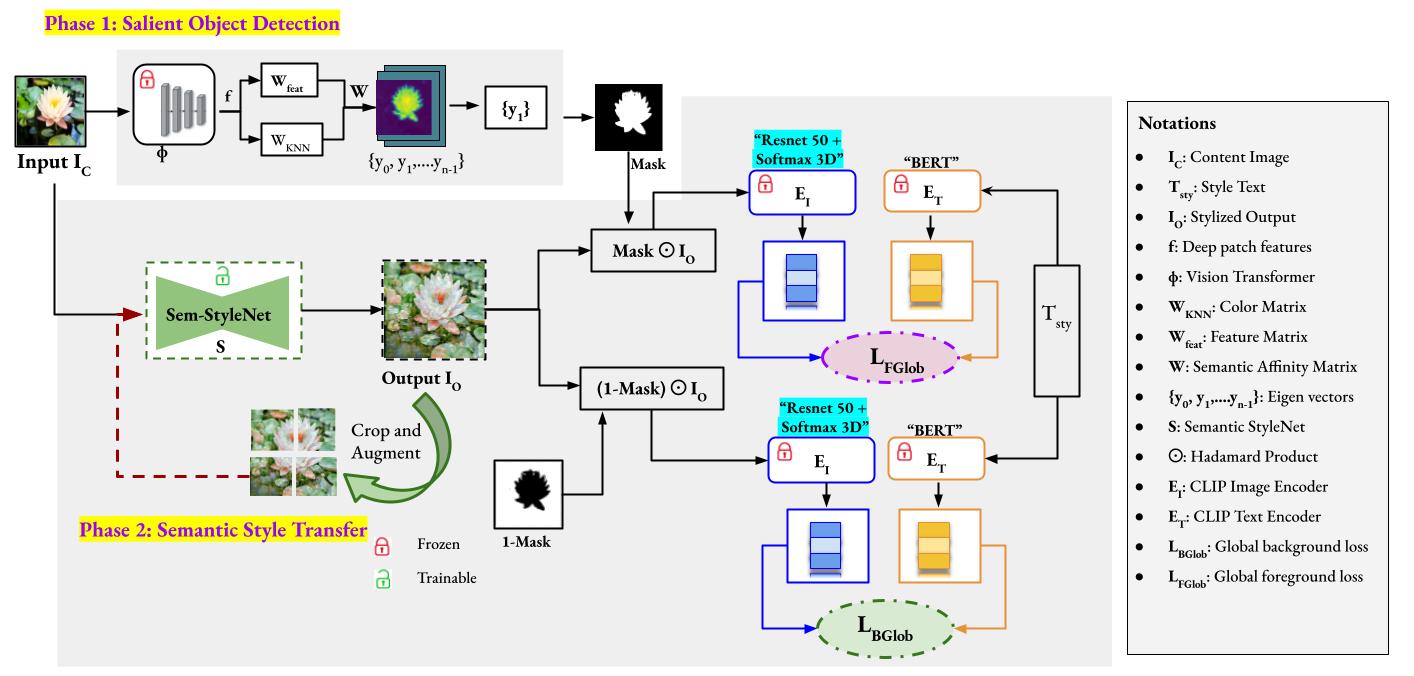}
        \vspace{-0.2cm}
    \caption{\small{The figure shows Semantic CLIPStyler (Sem-CS) framework. The two phases of Sem-CS are Salient Object Detection and Semantic Style Transfer are shown at the top \& bottom. The proposed \textit{global foreground} and \textit{background loss} are illustrated in the middle-right.}}
        \label{fig:block1}
    \end{figure*}
    
Generative Artisan (Gen-Art)~\cite{yang2022generative} addresses the over-styling problem of CLIPStyler~\cite{Kwon_2022_CVPR} through an FCN semantic segmentation network~\cite{long2015fully}. They control the degree of image style transfer in different semantic chunks. However, the supervised approach to extract the semantic parts of the content image needs to be more generalizable. \textit{E.g.}, their FCN semantic segmentation network only considers 21 classes; this is insufficient to represent real-world images. Also, they do not address content mismatch (see Fig.~\ref{fig:main_fig}).

In this paper, we propose Semantic CLIPStyler (Sem-CS), which addresses the content mismatch and over-styling problems of text-condition based style transfer. We use the deep spectral segmentation network~\cite{melas2022deep}, which extracts salient and non-salient objects of the content image in an unsupervised manner. As such, our method is generalizes well for real-world images. 

Sem-CS applies styles on salient or non-salient objects based on the text conditions. The key idea is to perform semantic style transfer using the proposed \textit{global foreground} and \textit{background loss}. Sem-CS also achieves controllable generation in transferring texture information for multiple text conditions. Our major contributions are as follows:
    \begin{itemize}[noitemsep,leftmargin=*]
        \item We propose a novel framework (Sem-CS) to perform style transfer with a text condition (Algorithm~\ref{alg:cap}). 
        \item We propose global foreground and global background loss to supervise style features semantically on the output (Sec.~\ref{sec:approach}).  
        \item We provide a reference-based quality assessment using DISTS~\cite{ding2020image} as well as a no-reference quality assessment using NIMA~\cite{talebi2018nima} to show Sem-CS outperforms baselines (Table~\ref{table:nima_dist}). 
    \end{itemize}
\vspace{-0.3cm}
     \begin{algorithm}[!h]
    \setstretch{1.08}
    \caption{Semantic CLIPStyler framework.}
    \label{alg:cap}
    \begin{algorithmic}[1] \vspace{3pt}
    \State \textbf{\textsc{Sem-CS}($E_T$, $E_I$, $I_C$, $T_{sty}$, $\phi$, $S$)}
    \LeftComment {{\color{gray} \textit{Compute Mask for salient objects identification}}} 
    \State \hspace{0.5cm}  $W$ = AffinityMatrix($I_c$, $\phi$, )
    \State \hspace{0.5cm}  $\{y_0$, $y_1$, $\ldots$, $y_{n-1}\}$ = Eigen\_Decomposition($W$)  
    \State \hspace{0.5cm}  $Mask$ =  Extract\_Salient\_Object($y_1$)  \vspace{4pt}
    \LeftComment {{\color{gray} \textit{Perform Semantic Style Transfer}}} 
    \State\hspace{0.5cm} $t_{fg}$, $t_{bg}$ = Parse\_Style\_Text($T_{sty}$)
    \State\hspace{0.5cm} $I_{fg}$, $I_{bg}$ = $Mask \odot S(I_C), (1-Mask) \odot S(I_C)$
    \LeftComment {{\color{teal} \textit{~~~~Global Foreground Loss}}}
    \State \hspace{0.5cm}  Compute Foreground Image Direction Loss $\Delta fg_{I}$  
    \State \hspace{0.5cm}  Compute Foreground Text Direction Loss $\Delta fg_{T}$
    \State \hspace{0.5cm} $\mathcal{L}_{FGlob}$ = Cosine\_similarity($\Delta fg_{I}, \Delta fg_{T}$)
    \LeftComment {{\color{teal} \textit{~~~~Global Background Loss}}}
    \State \hspace{0.5cm}  Compute Background Image Direction Loss $\Delta bg_{I}$ 
    \State \hspace{0.5cm} Compute Background Text Direction Loss $\Delta bg_{T}$
    \State \hspace{0.5cm} $\mathcal{L}_{BGlob}$ = Cosine\_similarity($\Delta bg_{I}$, $\Delta bg_{T}$)
    \LeftComment {{\color{teal} \textit{~~~~Minimize loss and compute output $I_O$}}}
    \State \hspace{0.5cm}  $I_O = \displaystyle\min_{{\theta_S}} \big( \mathcal{L}_{FGlob} + \lambda_{bg} \mathcal{L}_{BGlob} \big)$ 
\end{algorithmic}
\end{algorithm}

  \begin{figure*}[!h]
        \begin{minipage}{0.10\textwidth}
         \centering
            \small
            Input Image
        \end{minipage}
        \begin{minipage}{0.10\textwidth}
         \centering
         \small
            Style Text
        \end{minipage}
        \hfill
        \begin{minipage}{0.09\textwidth}
         \centering
         \small
            CLIP-Styler~\cite{Kwon_2022_CVPR}
        \end{minipage}
        \begin{minipage}{0.09\textwidth}
         \centering
         \small
            Gen-Art~\cite{yang2022generative}
        \end{minipage}
        \begin{minipage}{0.09\textwidth}
         \centering
         \small
            Sem-CS (ours)
        \end{minipage} 
        \hfill
        \hfill
        \begin{minipage}{0.10\textwidth}
         \centering
         \small
            Input Image
        \end{minipage}
        \begin{minipage}{0.10\textwidth}
         \centering
         \small
            Style text
        \end{minipage}
        \hfill
        \begin{minipage}{0.09\textwidth}
         \centering
         \small
            CLIP-Styler~\cite{Kwon_2022_CVPR}
        \end{minipage}
        \begin{minipage}{0.09\textwidth}
         \centering
         \small
            Gen-Art~\cite{yang2022generative}
        \end{minipage}
        \begin{minipage}{0.09\textwidth}
         \centering
         \small
            Sem-CS (ours)
        \end{minipage} 
       \begin{minipage}{0.09\textwidth}
         \centering
             \includegraphics[width=0.99\linewidth]{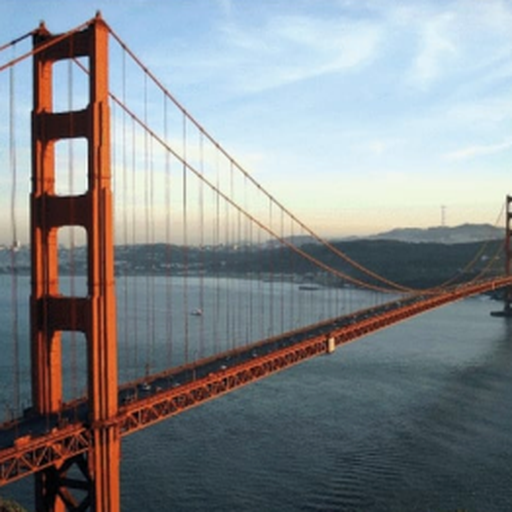}
        \end{minipage}
         \fbox{\begin{minipage}[c][1.35cm][c]{1.5cm}
         \centering
         \footnotesize
            Acrylic painting
        \end{minipage}}
        \hfill
        \begin{minipage}{0.09\textwidth}
         \centering
             \includegraphics[width=0.99\linewidth]{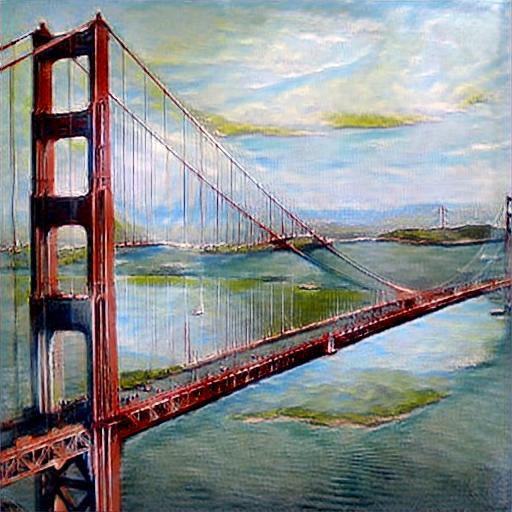}
        \end{minipage}
        \begin{minipage}{0.09\textwidth}
         \centering                 
            \includegraphics[width=0.99\linewidth]{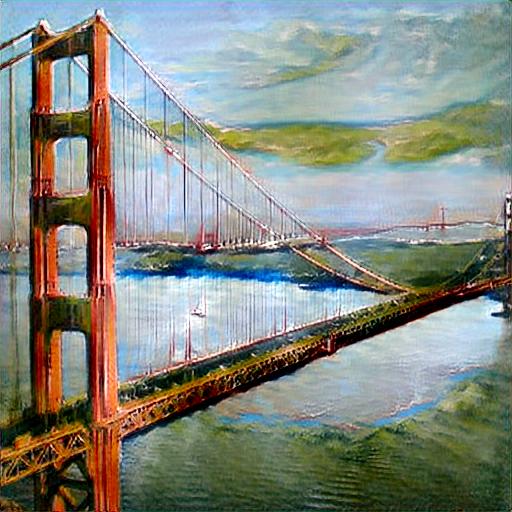}
        \end{minipage}
        \begin{minipage}{0.09\textwidth}
         \centering                 
            \includegraphics[width=0.99\linewidth]{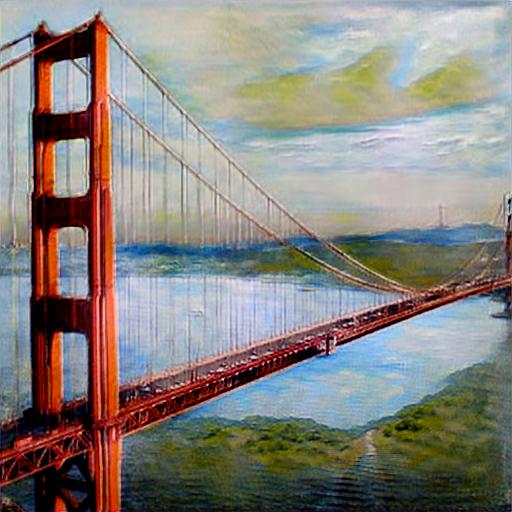}
        \end{minipage}
        \hfill
        \textbf{\vline}
        \hfill
    \begin{minipage}{0.09\textwidth}
         \centering             
         \includegraphics[width=0.99\linewidth]{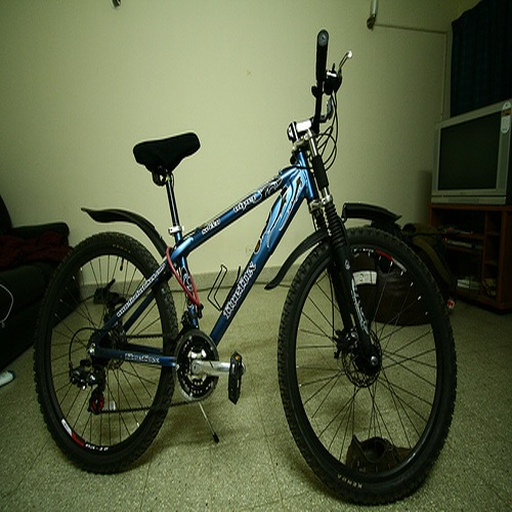}
        \end{minipage}
        \fbox{\begin{minipage}[c][1.35cm][c]{1.5cm}
         \centering
         \footnotesize
            Snowy
        \end{minipage}}
        \hfill
        \begin{minipage}{0.09\textwidth}
         \centering
             \includegraphics[width=0.99\linewidth]{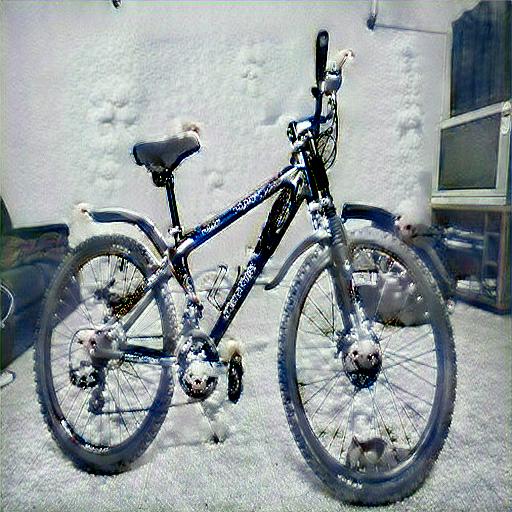}
        \end{minipage}
        \begin{minipage}{0.09\textwidth}
         \centering                 
            \includegraphics[width=0.99\linewidth]{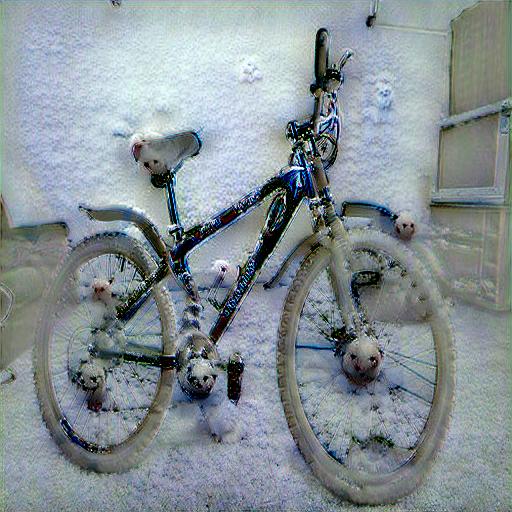}
        \end{minipage}
        \begin{minipage}{0.09\textwidth}
         \centering                 
            \includegraphics[width=0.99\linewidth]{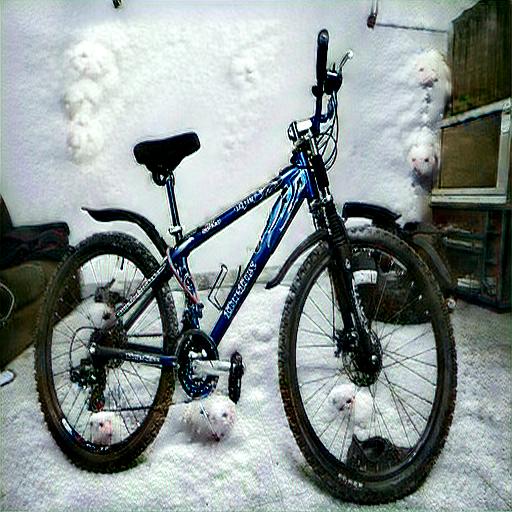}
        \end{minipage}
  
    \begin{minipage}{0.09\textwidth}
         \centering
             \includegraphics[width=0.99\linewidth]{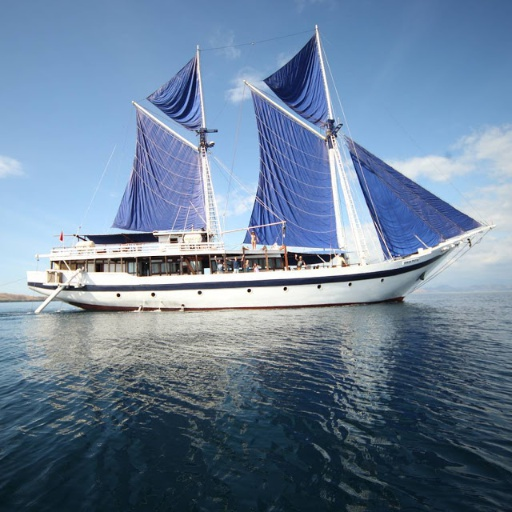}
        \end{minipage}
       \fbox{\begin{minipage}[c][1.35cm][c]{1.5cm}
         \centering
         \footnotesize
            A graffiti style painting
        \end{minipage}}
        \hfill
        \begin{minipage}{0.09\textwidth}
         \centering            
         \includegraphics[width=0.99\linewidth]{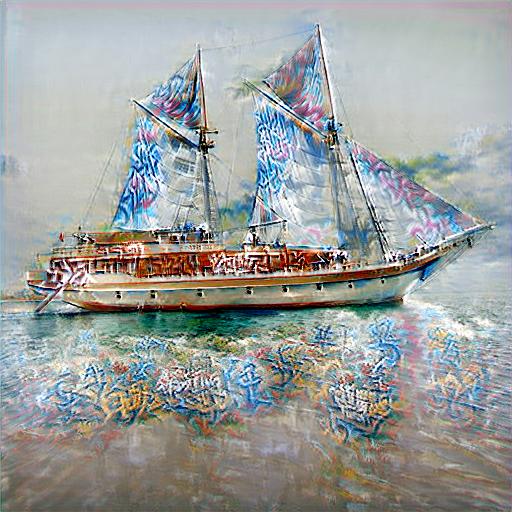}
        \end{minipage}
        \begin{minipage}{0.09\textwidth}
         \centering                 
        \includegraphics[width=0.99\linewidth]{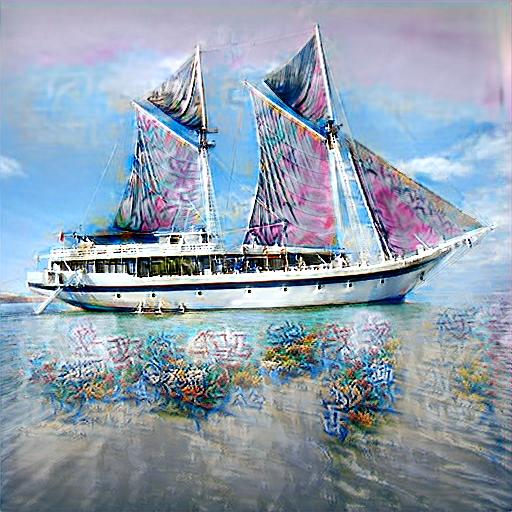}
        \end{minipage}
        \begin{minipage}{0.09\textwidth}
         \centering             
         \includegraphics[width=0.99\linewidth]{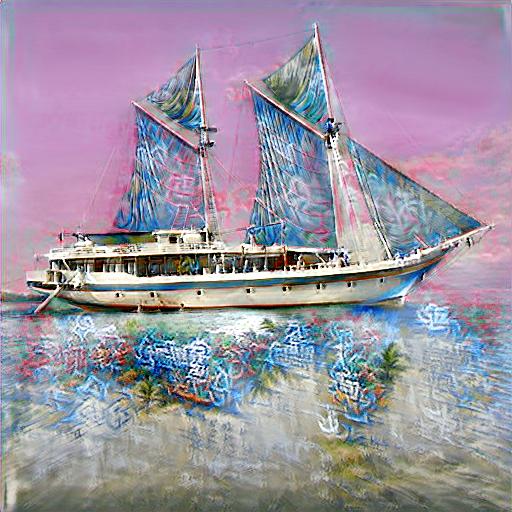}
        \end{minipage}
        \hfill
        \textbf{\vline}
        \hfill
        \begin{minipage}{0.09\textwidth}
         \centering
             \includegraphics[width=0.99\linewidth]{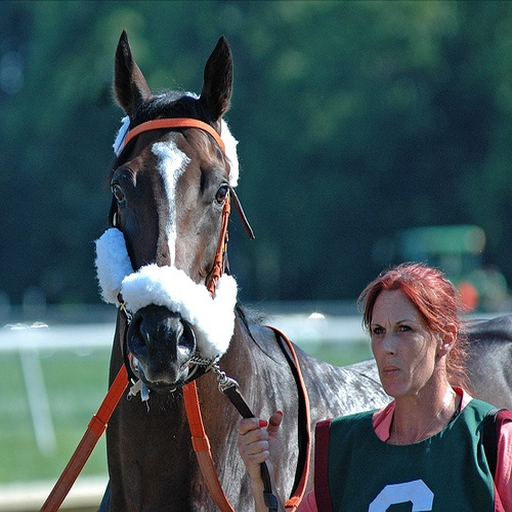}
        \end{minipage}
        \fbox{\begin{minipage}[c][1.35cm][c]{1.5cm}
         \centering
         \footnotesize
            Red rocks
        \end{minipage}}
        \hfill
        \begin{minipage}{0.09\textwidth}
         \centering
             \includegraphics[width=0.99\linewidth]{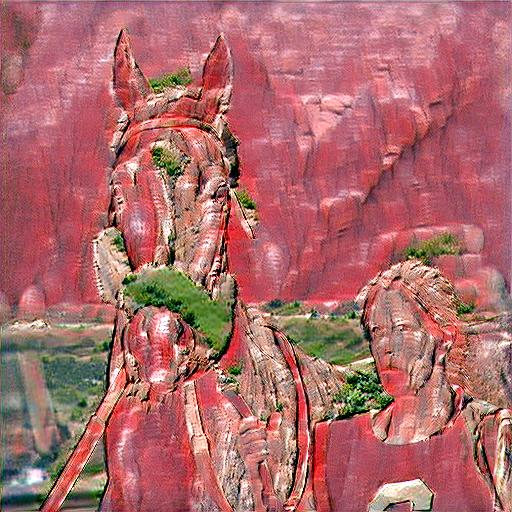}
        \end{minipage}
        \begin{minipage}{0.09\textwidth}
         \centering                 
            \includegraphics[width=0.99\linewidth]{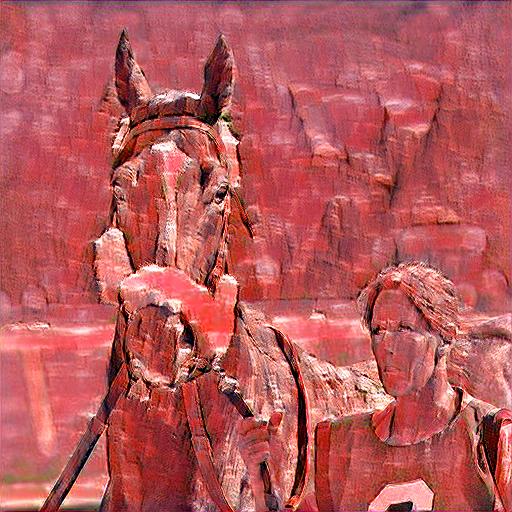}
        \end{minipage}
        \begin{minipage}{0.09\textwidth}
         \centering                 
            \includegraphics[width=0.99\linewidth]{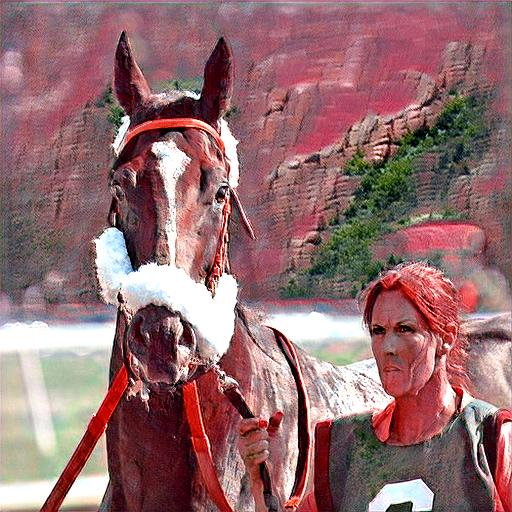}
        \end{minipage}

                \begin{minipage}{0.09\textwidth}
         \centering
             \includegraphics[width=0.99\linewidth]{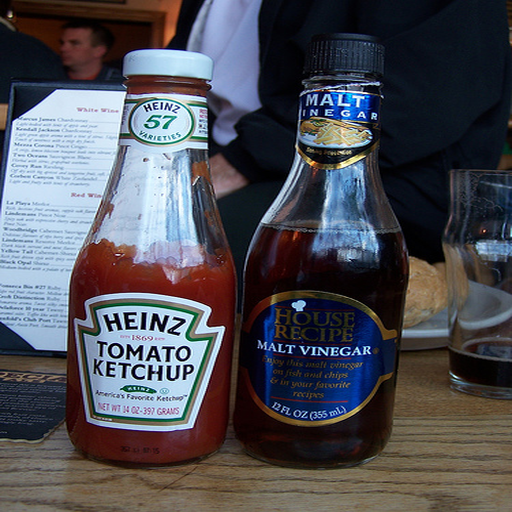}
        \end{minipage}
        \fbox{\begin{minipage}[c][1.35cm][c]{1.5cm}
         \centering
         \footnotesize
            A fauvism style painting
        \end{minipage}}
        \hfill
        \begin{minipage}{0.09\textwidth}
         \centering             
         \includegraphics[width=0.99\linewidth]{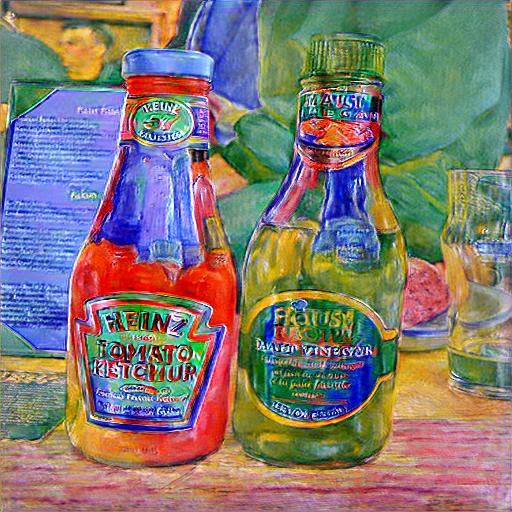}
        \end{minipage}
        \begin{minipage}{0.09\textwidth}
         \centering            
        \includegraphics[width=0.99\linewidth]{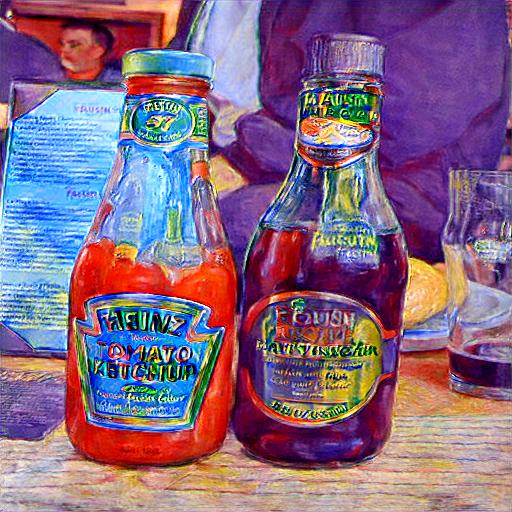}
        \end{minipage}
        \begin{minipage}{0.09\textwidth}
         \centering                 
        \includegraphics[width=0.99\linewidth]{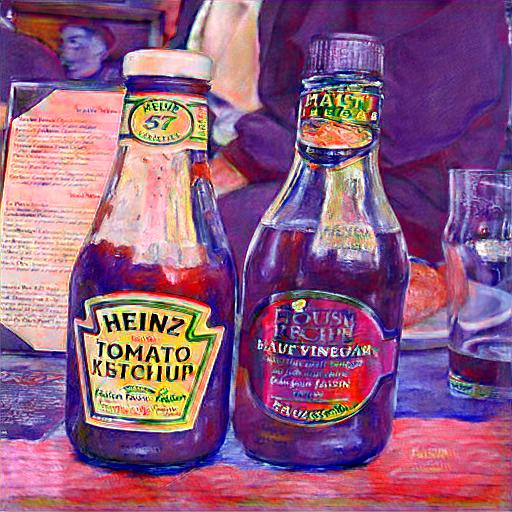}
        \end{minipage}
        \hfill
        \textbf{\vline}
        \hfill
        \begin{minipage}{0.09\textwidth}
         \centering
             \includegraphics[width=0.99\linewidth]{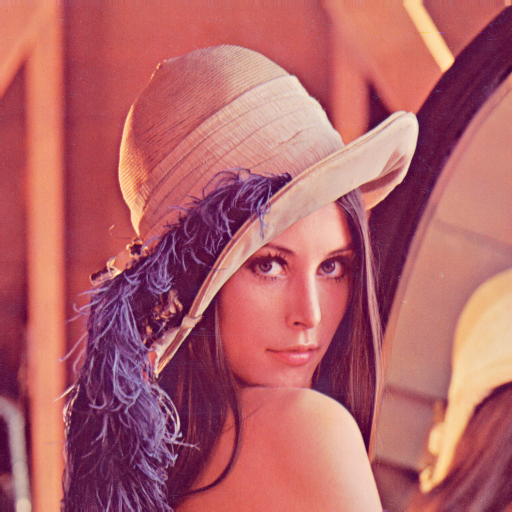}
        \end{minipage}
        \fbox{\begin{minipage}[c][1.35cm][c]{1.55cm}
         \centering
         \footnotesize
            A watercolor painting with purple brush
        \end{minipage}}
        \hfill
        \begin{minipage}{0.09\textwidth}
         \centering
             \includegraphics[width=0.99\linewidth]{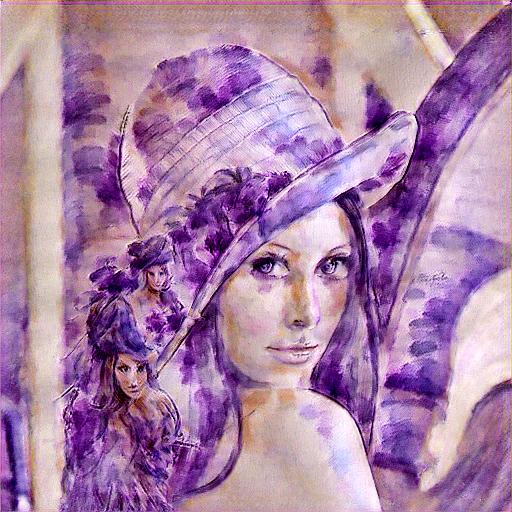}
        \end{minipage}
        \begin{minipage}{0.09\textwidth}
         \centering                 
            \includegraphics[width=0.99\linewidth]{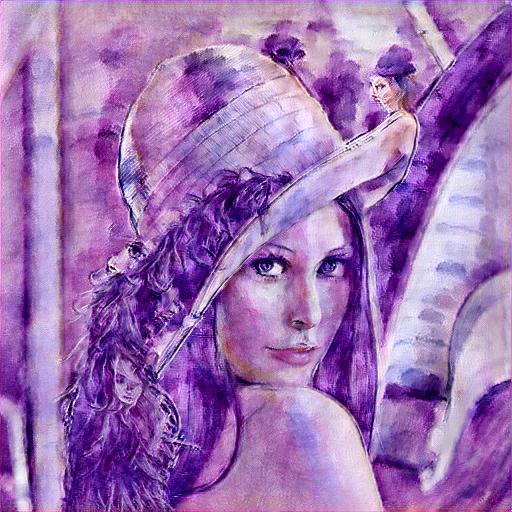}
        \end{minipage}
        \begin{minipage}{0.09\textwidth}
         \centering                 
            \includegraphics[width=0.99\linewidth]{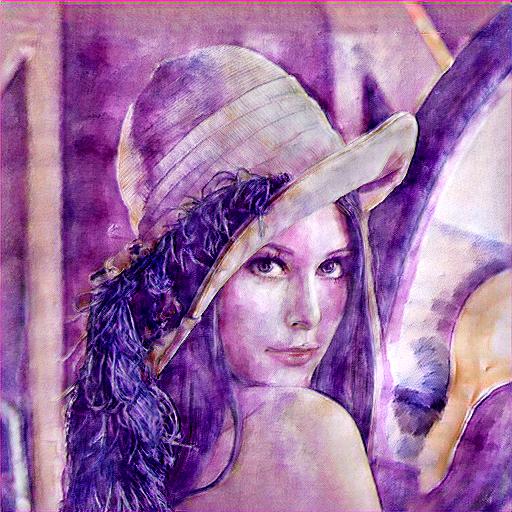}
        \end{minipage}
                \begin{minipage}{0.09\textwidth}
         \centering
             \includegraphics[width=0.99\linewidth]{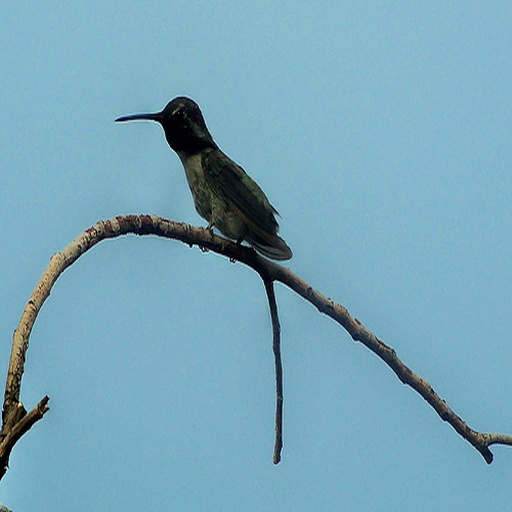}
        \end{minipage}
        \fbox{\begin{minipage}[c][1.35cm][c]{1.5cm}
         \centering
         \footnotesize
            An oil painting of white roses
        \end{minipage}}
        \hfill
        \begin{minipage}{0.09\textwidth}
         \centering             
         \includegraphics[width=0.99\linewidth]{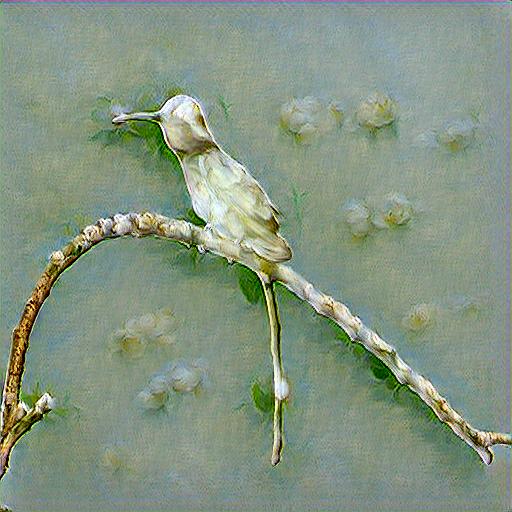}
        \end{minipage}
        \begin{minipage}{0.09\textwidth}
         \centering            
        \includegraphics[width=0.99\linewidth]{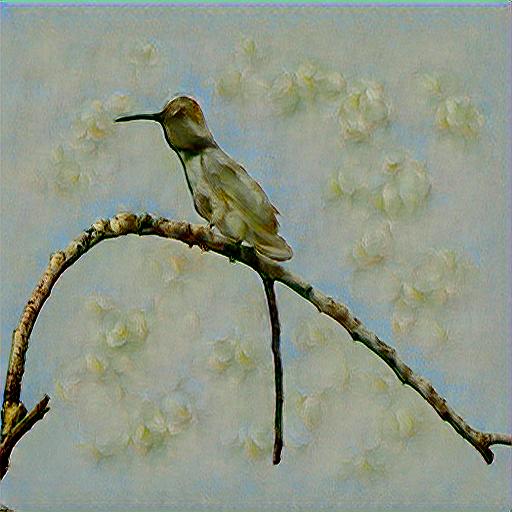}
        \end{minipage}
        \begin{minipage}{0.09\textwidth}
         \centering                 
        \includegraphics[width=0.99\linewidth]{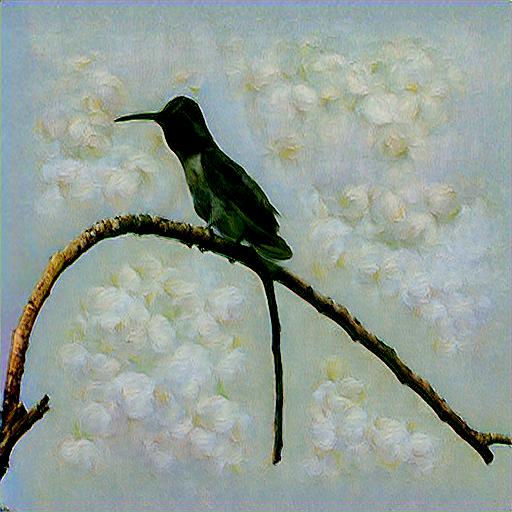}
        \end{minipage}
        \hfill
        \textbf{\vline}
        \hfill
        \begin{minipage}{0.09\textwidth}
         \centering
             \includegraphics[width=0.99\linewidth]{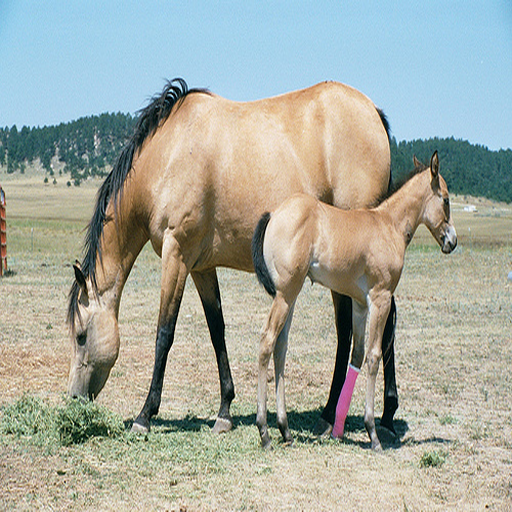}
        \end{minipage}
        \fbox{\begin{minipage}[c][1.35cm][c]{1.55cm}
         \centering
         \footnotesize
            A watercolor painting of leaf
        \end{minipage}}
        \hfill
        \begin{minipage}{0.09\textwidth}
         \centering
             \includegraphics[width=0.99\linewidth]{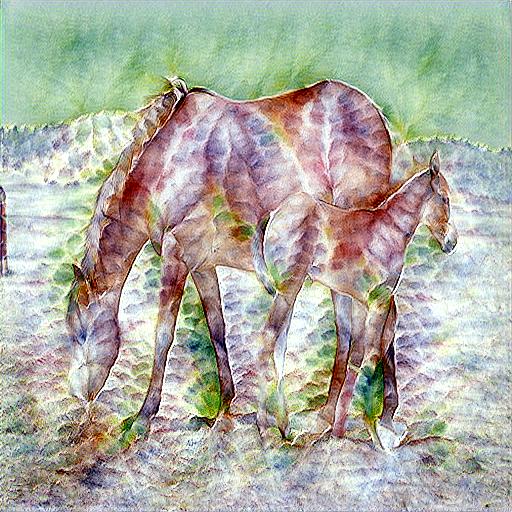}
        \end{minipage}
        \begin{minipage}{0.09\textwidth}
         \centering                 
            \includegraphics[width=0.99\linewidth]{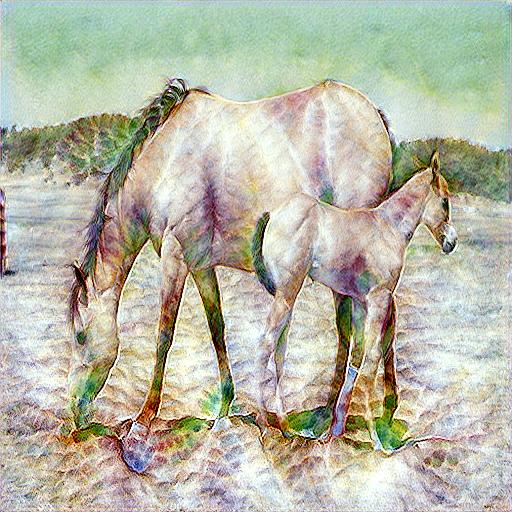}
        \end{minipage}
        \begin{minipage}{0.09\textwidth}
         \centering              
    \includegraphics[width=0.99\linewidth]{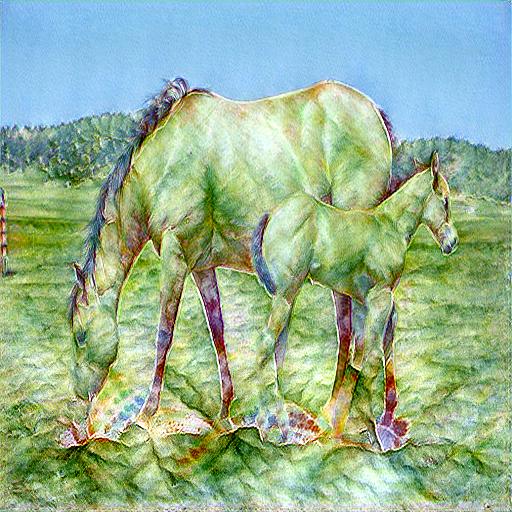}
        \end{minipage}
    \vspace{-0.1cm}
    \caption{\small{The figure shows the visual comparison of style transfer outputs with single text condition a) The text-based image style transfer outputs on the left shows that CLIPStyler~\cite{Kwon_2022_CVPR} and Gen-Art~\cite{yang2022generative} suffers from over-stylization. Sem-CS (ours) reduces the effects of over-stylization. b) Similarly, the outputs on the right shows that baseline methods suffers from content mismatch problem. Sem-CS (ours)  reduces content mismatch problem {\color{blue} (images are best viewed after zooming)}.}}
    \label{fig:single_style_res}
\end{figure*}

\section{Our Method}
\label{sec:approach}
This section describes our framework. It has two major phases: Salient Object Detection and Semantic Style Transfer. We illustrate Sem-CS in Fig.~\ref{fig:block1} and Algorithm~\ref{alg:cap} formally describes the proposed framework. The two phases of Sem-CS are described as follows. \vspace{2pt}

     \noindent \textit{\textbf{Salient Object Detection:}} In the first phase, we compute the masks for salient objects in the content image; see Algorithm~\ref{alg:cap}, lines 2-4. (The mask for salient objects is computed in an unsupervised setting.) First, we compute the affinity matrix (W) of the content image $I_C$ from the attention block of the last layers of feature extractor $\phi$. Secondly, we find the eigenvectors of the laplacian of the affinity matrix. Finally, we extract the mask from the eigenvector $y_1$. \vspace{2pt}
     
     \noindent \textit{\textbf{Semantic Style Transfer:}} In the second phase, we train Sem-StyleNet $S$ to transfer style features to the salient objects and background objects based on the text conditions (Fig.~\ref{fig:block1}). We use ResNet50 with softmax3d~\cite{wang2021rethinking} for the image encoder to make the stylized output more robust. We propose \textit{global foreground loss} and \textit{global background loss} for style supervision on salient objects and the background of the output image, respectively. These are:

     \smallskip



    \noindent \textbf{\textit{Global Foreground Loss}.} This ensures that relevant style text applies to the salient objects present in the output. To maintain the diversity of generated stylized outputs, directional CLIP loss~\cite{gal2022stylegan} is computed instead of global CLIP loss~\cite{patashnik2021styleclip} by aligning the CLIP-space direction between the text-image pairs of input and output. Foreground text directional loss  $(\Delta fg_{T})$ is defined to be the difference between source text embedding $(t_{src})$ and foreground style text embedding $(t_{fg})$ as described in Eq.~\ref{eq:fg_t}.   
         \begin{equation}
            \label{eq:fg_t}
            \Delta fg_{T} = E_{T}(t_{fg}) - E_{T}(t_{src})
    \end{equation}
    Here, $E_{T}$ is the CLIP text-encoder and $t_{src}$ is set to "Photo". Foreground image directional loss $(\Delta fg_{I})$ is computed between embeddings of salient objects and style transfer output. Given the content image $I_C$ and $Mask$, Hadamard product $\odot$ is computed between $Mask$ and $S(I_C)$ to extract features for salient objects as $I_{fg} = Mask \odot  S(I_C)$. Next, $\Delta fg_{I}$ is computed as described in Eq.~\ref{eq:fg_i}.
        \begin{equation}
        \label{eq:fg_i}
        \Delta fg_{I} = E_{I}(I_{fg}) - E_I(I_C)
    \end{equation}
    $E_{I}$ is the CLIP image encoder. Finally, Global foreground loss $(\mathcal{L}_{FGlob})$ is computed by taking cosine similarity between CLIP-Space direction of the foreground of image and style texts (Eq.~\ref{eq:fg_glob}). 
    \begin{equation}
        \label{eq:fg_glob}
        \mathcal{L}_{FGlob} = 1- \frac{\Delta fg_{I} . \Delta fg_{T}}{\lvert{\Delta fg_{I}}\rvert \lvert{\Delta fg_{T}}\rvert}
    \end{equation}
    Here, one minus the cosine similarity represents the distance between image and text directional loss. In other words, the global foreground loss minimizes the distance between the image direction loss and text direction loss for salient objects.

    \smallskip
   
    \noindent \textbf{\textit{Global Background Loss}.} This is computed for style feature supervision of the output image background. Similar to global foreground loss, we compute background text directional loss $(\Delta bg_{T})$ for style background as given in Eq.~\ref{eq:bg_T}.
         \begin{equation}
         \label{eq:bg_T}
            \Delta bg_{T} = E_{T}(t_{bg}) - E_{T}(t_{src})
    \end{equation}
    Here, $t_{bg}$ is the style text condition for the background. Also, background image directional loss $\Delta bg_{I}$ is computed as shown in Eq.~\ref{eq:bg_i}. We take Hadamard product between the background mask and generated image $I_{bg} = (1-Mask) \odot I_O$ to extract background features. Next, $\Delta bg_{I}$ is computed as below in Eq.~\ref{eq:bg_i}
       \vspace{-0.2cm}
        \begin{equation}
        \label{eq:bg_i}
        \Delta bg_{I} = E_{I}(I_{bg}) - E_I(I_C)
    \end{equation}
    Finally, global background loss $\mathcal{L}_{BGlob}$ is computed to minimize the distance between image and text directional losses for background objects as described in Eq.~\ref{eq:bg_loss}.
    \begin{equation}
    \label{eq:bg_loss}
        \mathcal{L}_{BGlob} = 1- \frac{\Delta bg_{I} . \Delta bg_{T}}{\lvert{\Delta bg_{I}}\rvert \lvert{\Delta bg_{T}}\rvert}
    \end{equation}   
    Here, global background loss $\mathcal{L}_{BGlob}$ helps to perform controllable style transfer for background objects in the style transfer outputs. \vspace{2pt}

    \noindent \textbf{\textit{Other Loss}.} We also add content loss and a total variation regularization loss to our proposed loss for style transfer \cite{Kwon_2022_CVPR}.

\begin{table}[!ht]\small{
\begin{center}\renewcommand{\arraystretch}{1.25}
\begin{tabular}{cccc}
                                         & \cellcolor[HTML]{ECF4FF}CLIPStyler~\cite{Kwon_2022_CVPR} & \cellcolor[HTML]{EFEFEF}Gen-Art~\cite{yang2022generative} & \cellcolor[HTML]{FFFFC7}Sem-CS (ours) \\ \cline{2-4} 
\multicolumn{1}{l|}{\textit{DISTS}~\cite{ding2020image}}      & \multicolumn{1}{c|}{0.32}              & \multicolumn{1}{c|}{0.25}                      & \multicolumn{1}{c|}{\textbf{0.34}}                 \\ \cline{2-4} 
\multicolumn{1}{l|}{\textit{NIMA}~\cite{talebi2018nima}}       & \multicolumn{1}{c|}{4.61}              & \multicolumn{1}{c|}{4.34}                      & \multicolumn{1}{c|}{\textbf{5.34}}                 \\ \cline{2-4} 
\multicolumn{1}{l|}{\textit{User Study}} & \multicolumn{1}{c|}{28.3}              & \multicolumn{1}{c|}{33.1}                      & \multicolumn{1}{c|}{\textbf{38.4}}                 \\ \cline{2-4} 
\end{tabular}\end{center}}
\caption{\small{The table shows quantitative (DISTS~\cite{ding2020image} and NIMA~\cite{talebi2018nima} scores) and qualitative (User Study) evaluations.}}
\label{table:nima_dist}
\end{table}
\section{Experimental Results}
\label{ssec:subheadresult}

Fig.~\ref{fig:single_style_res} shows that Sem-CS preserves the semantics of objects in output images while minimizing over-stylization and content mismatch. For example, let us see first row on left side of Fig.~\ref{fig:single_style_res}. It could be observed that CLIPStyler~\cite{Kwon_2022_CVPR} and Generative Artisan~\cite{yang2022generative} outputs are over-stylized (the "\textit{Acrylic}" style spills both below the bridge and onto the sky), and the content features of the water are lost. Sem-CS (ours) preserves the semantics of the bridge. Similarly, in the first row, on the right side, CLIPStyler~\cite{Kwon_2022_CVPR} and Generative Artisan~\cite{yang2022generative} outputs suffer from content mismatch as the \textit{"Snowy"} style is applied to the bicycle and background. Sem-CS performs style transfer while minimizing the content mismatch effects of the \textit{"Snowy"} style feature.

We evaluated Sem-CS framework with DISTS~\cite{ding2020image}, NIMA~\cite{talebi2018nima}, and a User Study (Table~\ref{table:nima_dist}). We describe the quantitative results as follows. 

\noindent \textbf{\textit{DISTS~\cite{ding2020image} Scores.}} DISTS~\cite{ding2020image} is a  reference-based image quality assessment that shows the preservation of object structure in the presence of texture transfer in stylized output (how well they are preserved); since DISTS~\cite{ding2020image} may not capture all aspects of style transfer quality like semantic coherence, we add NIMA~\cite{talebi2018nima} scores to support it and also conduct a user study Table~\ref{table:nima_dist}.

\noindent \textbf{\textit{NIMA~\cite{talebi2018nima} Scores.}} NIMA~\cite{talebi2018nima} is a no-reference-based image quality metric that predicts the quality of distribution ratings with a significant correlation to ground truth ratings.
Table~\ref{table:nima_dist} reports the average scores of top-100 output images. 
        

    \vspace{2pt}
    \noindent \textbf{\textit{User Study.}}
     We conducted a user study to validate preserved semantics of objects while transferring the style texts onto the content image (Table~\ref{table:nima_dist}). We randomly sampled 5 groups of 15 images from the outputs produced above, with ten images from single-style text and five images from double-style text stylized outputs. All 5 $\times$ 15 stylized outputs were distributed anonymously and randomly to 40 participants. They were asked to observe the stylized results from different methods and vote for the image that looks better in quality and matches the style text. 
     Table~\ref{table:nima_dist} shows the percentage vote for each method. Sem-CS outperforms baseline methods. \vspace{2pt}   

Overall, we find that Sem-CS scores are higher than the baseline methods CLIPStyler~\cite{Kwon_2022_CVPR} and Generative Artisan~\cite{yang2022generative}. This justifies that adding global foreground and background losses improves the image quality of stylized output. Sem-CS minimizes content mismatch and prevents distortion of objects present in output image when supervising style features. 


\begin{figure}[!htb]
       \centering
        \begin{minipage}{0.25\linewidth}
         \centering
         \small
            \textbf{Input Image}
        \end{minipage}
      \begin{minipage}{0.25\linewidth}
         \centering
         \small
            Gen-Art~\cite{yang2022generative}
        \end{minipage}
      \begin{minipage}{0.25\linewidth}
         \centering
         \small
            Sem-CS (ours)
        \end{minipage}
    \begin{minipage}{0.25\linewidth}
         \centering
             \includegraphics[width=0.99\linewidth]{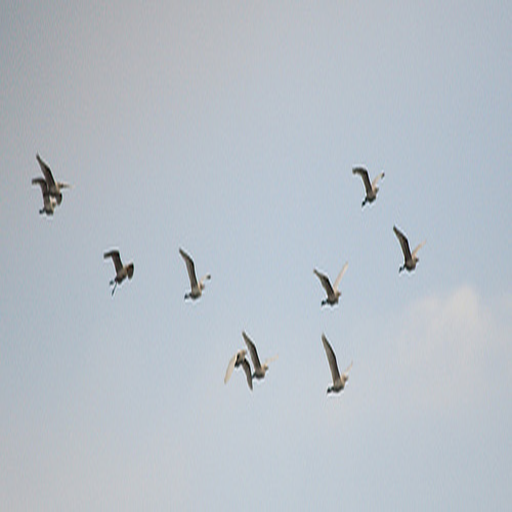}
        \end{minipage}
        \begin{minipage}{0.25\linewidth}
         \centering
             \includegraphics[width=0.99\linewidth]{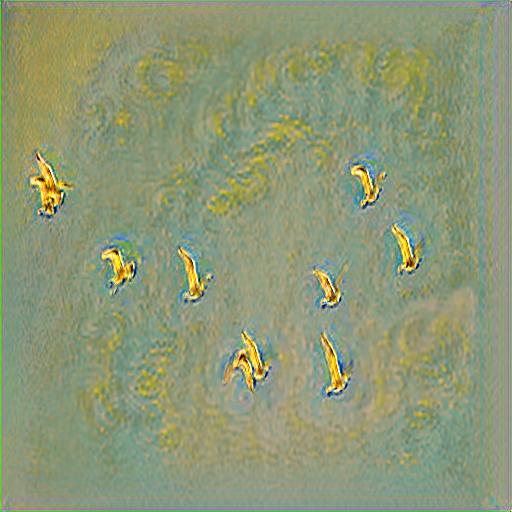}
        \end{minipage}
        \begin{minipage}{0.25\linewidth}
         \centering
             \includegraphics[width=0.99\linewidth]{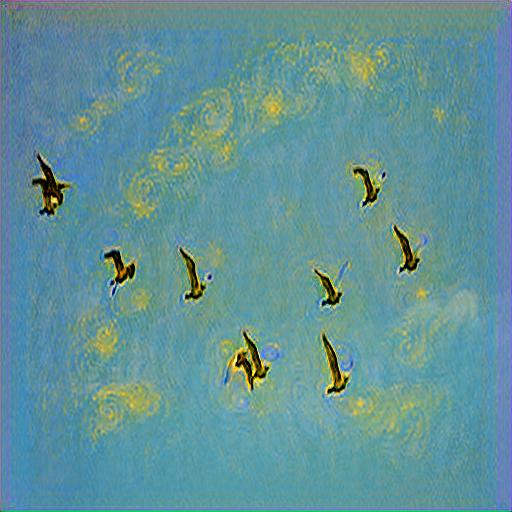}
        \end{minipage}
        
        \begin{minipage}{0.25\linewidth}
         \centering
             \includegraphics[width=0.99\linewidth]{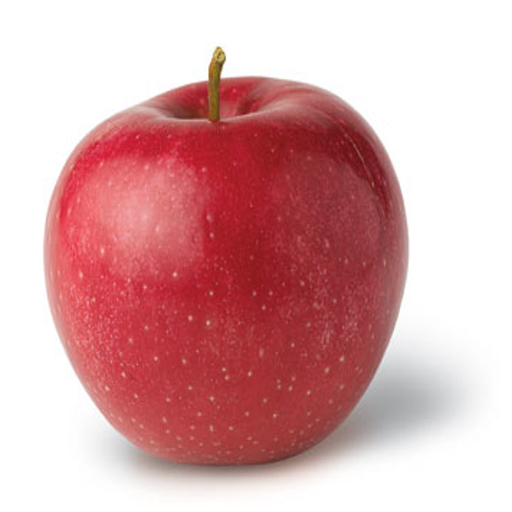}
        \end{minipage}
        \begin{minipage}{0.25\linewidth}
         \centering
             \includegraphics[width=0.99\linewidth]{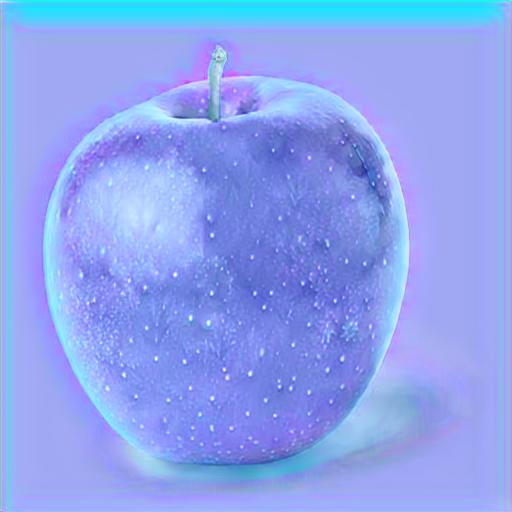}
        \end{minipage}
        \begin{minipage}{0.25\linewidth}
         \centering
             \includegraphics[width=0.99\linewidth]{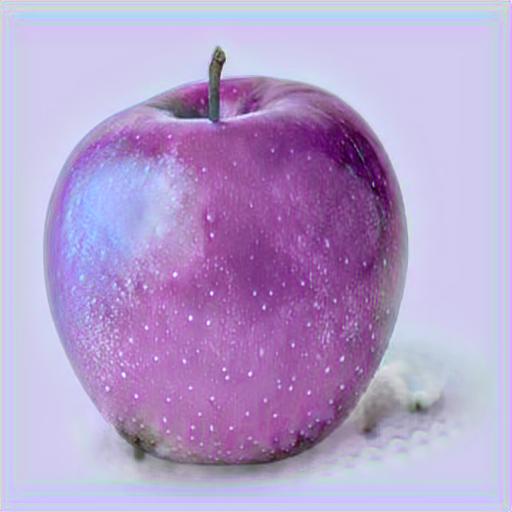}
        \end{minipage}
    \vspace{-0.1cm}
        \caption{\small{Stylized outputs with two text conditions (foreground \textbf{F} and  background \textbf{B}). a) Top Row: \textbf{F}: Pop Art; \textbf{B}: Starry Night by Vincent Van Gogh. b) Bottom Row: \textbf{F}: Red Rocks; \textbf{B}: Snowy.}}
        \label{fig:double_style_res}
    \end{figure}
    
\noindent \textbf{\textit{Ablation Studies.}} Fig.~\ref{fig:double_style_res} illustrate ablation studies for style transfer using double style texts condition. The double-style texts are challenging because style supervision is required for salient objects and backgrounds of image. Therefore, double-style texts require more controllable generation capabilities for style transfer. We evaluated Sem-CS framework for double style-texts with  NIMA~\cite{talebi2018nima} and  DISTS~\cite{ding2020image} scores on 100 stylized outputs. Table~\ref{table:ablation} describes that Sem-CS outperforms Generative Artisan~\cite{yang2022generative}. Also, note that the user study scores of style transfer outputs for double text condition for Sem-CS are higher. 
\begin{table}[!ht]\small{
\begin{center}\renewcommand{\arraystretch}{1.25}
\begin{tabular}{ccc}
                                         & \cellcolor[HTML]{EFEFEF}Gen-Art \cite{yang2022generative} & \cellcolor[HTML]{FFFFC7}Sem-CS (ours) \\ \cline{2-3} 
\multicolumn{1}{l|}{\textit{DISTS}~\cite{ding2020image}}      & \multicolumn{1}{c|}{0.20}                      & \multicolumn{1}{c|}{\textbf{0.33}}                 \\ \cline{2-3} 
\multicolumn{1}{l|}{\textit{NIMA}~\cite{talebi2018nima}}       & \multicolumn{1}{c|}{4.34}                      & \multicolumn{1}{c|}{\textbf{5.52}}                 \\ \cline{2-3} 
\multicolumn{1}{l|}{\textit{User Study}} & \multicolumn{1}{c|}{48.2}                      & \multicolumn{1}{c|}{\textbf{51.7}}                 \\ \cline{2-3} 
\end{tabular}\end{center}}
\vspace{-0.4cm}
\caption{\small{\textbf{Ablation Study.} Scores for an ablation study performed for controllable generation using multiple text conditions.}}
\label{table:ablation}
\end{table}
\section{Conclusion}
We proposed Semantic CLIPStyler (Sem-CS) to preserve the semantics of objects and prevent over-stylization when performing text-based image style transfer. We showed that style transfer could be done semantically by training the StyleNet with the proposed global background and foreground loss. Our quantitative and qualitative experimental results showed that Sem-CS achieves superior stylized output with text descriptions. The scope of future work extends to applying different text conditions on more than one object present in the content image. For this, we aim to improve the segmentation mask of content image.

\bibliographystyle{IEEEbib}

\bibliography{refs}

\begin{thebibliography}{10}

\bibitem{gatys2016image}
Leon~A Gatys, Alexander~S Ecker, and Matthias Bethge,
\newblock ``Image style transfer using convolutional neural networks,''
\newblock in {\em Proceedings of the IEEE conference on computer vision and
  pattern recognition}, 2016.

\bibitem{NEURIPS2021_df535469}
Haibo Chen, lei zhao, Zhizhong Wang, Huiming Zhang, Zhiwen Zuo, Ailin Li, Wei
  Xing, and Dongming Lu,
\newblock ``Artistic style transfer with internal-external learning and
  contrastive learning,''
\newblock in {\em Advances in Neural Information Processing Systems},
  M.~Ranzato, A.~Beygelzimer, Y.~Dauphin, P.S. Liang, and J.~Wortman Vaughan,
  Eds. 2021, Curran Associates, Inc.

\bibitem{li2017universal}
Yijun Li, Chen Fang, Jimei Yang, Zhaowen Wang, Xin Lu, and Ming-Hsuan Yang,
\newblock ``Universal style transfer via feature transforms,''
\newblock {\em Advances in neural information processing systems}, 2017.

\bibitem{park2019arbitrary}
Dae~Young Park and Kwang~Hee Lee,
\newblock ``Arbitrary style transfer with style-attentional networks,''
\newblock in {\em proceedings of the IEEE/CVF conference on computer vision and
  pattern recognition}, 2019.

\bibitem{mastan2022dilie}
Indra~Deep Mastan, Shanmuganathan Raman, and Prajwal Singh,
\newblock ``Dilie: Deep internal learning for image enhancement,''
\newblock in {\em Proceedings of the IEEE/CVF Winter Conference on Applications
  of Computer Vision}, 2022.

\bibitem{mechrez2018contextual}
Roey Mechrez, Itamar Talmi, and Lihi Zelnik{-}Manor,
\newblock ``The contextual loss for image transformation with non-aligned
  data,''
\newblock in {\em Computer Vision - {ECCV} 2018 - 15th European Conference,
  Munich, Germany, September 8-14, 2018, Proceedings, Part {XIV}}. 2018,
  Lecture Notes in Computer Science, Springer.

\bibitem{8451734}
Sahil Chelaramani, Abhishek Jha, and Anoop~M. Namboodiri,
\newblock ``Cross-modal style transfer,''
\newblock in {\em 2018 25th IEEE International Conference on Image Processing
  (ICIP)}, 2018.

\bibitem{BMVC2017_153}
Eli~Shechtman Roey~Mechrez and Lihi Zelnik-Manor,
\newblock ``Photorealistic style transfer with screened poisson equation,''
\newblock in {\em Proceedings of the British Machine Vision Conference (BMVC)}.
  2017, BMVA Press.

\bibitem{Luan2017DeepPS}
Fujun Luan, Sylvain Paris, Eli Shechtman, and Kavita Bala,
\newblock ``Deep photo style transfer,''
\newblock {\em 2017 IEEE Conference on Computer Vision and Pattern Recognition
  (CVPR)}, 2017.

\bibitem{samuth2022patch}
Benjamin Samuth, David Tschumperl{\'e}, and Julien Rabin,
\newblock ``A patch-based approach for artistic style transfer via constrained
  multi-scale image matching,''
\newblock in {\em 2022 IEEE International Conference on Image Processing
  (ICIP)}. IEEE, 2022.

\bibitem{Kwon_2022_CVPR}
Gihyun Kwon and Jong~Chul Ye,
\newblock ``Clipstyler: Image style transfer with a single text condition,''
\newblock in {\em Proceedings of the IEEE/CVF Conference on Computer Vision and
  Pattern Recognition (CVPR)}, June 2022.

\bibitem{yang2022generative}
Zhenling Yang, Huacheng Song, and Qiunan Wu,
\newblock ``Generative artisan: A semantic-aware and controllable clipstyler,''
\newblock {\em arXiv preprint arXiv:2207.11598}, 2022.

\bibitem{Mastan_2021_CVPR}
Indra~Deep Mastan and Shanmuganathan Raman,
\newblock ``Deepobjstyle: Deep object-based photo style transfer,''
\newblock in {\em Proceedings of the IEEE/CVF Conference on Computer Vision and
  Pattern Recognition (CVPR) Workshops}, June 2021.

\bibitem{context2022icvgip}
Chanda Grover, Indra~Deep Mastan, and Debayan Gupta,
\newblock ``Contextclip: Contextual alignment of image-text pairs on clip
  visual representations,''
\newblock in {\em Proceedings of the Indian Conference on Computer Vision,
  Graphics and Image Processing}, 2023.

\bibitem{long2015fully}
Jonathan Long, Evan Shelhamer, and Trevor Darrell,
\newblock ``Fully convolutional networks for semantic segmentation,''
\newblock in {\em Proceedings of the IEEE conference on computer vision and
  pattern recognition}, 2015.

\bibitem{melas2022deep}
Luke Melas-Kyriazi, Christian Rupprecht, Iro Laina, and Andrea Vedaldi,
\newblock ``Deep spectral methods: A surprisingly strong baseline for
  unsupervised semantic segmentation and localization,''
\newblock in {\em Proceedings of the IEEE/CVF Conference on Computer Vision and
  Pattern Recognition}, 2022.

\bibitem{ding2020image}
Keyan Ding, Kede Ma, Shiqi Wang, and Eero~P Simoncelli,
\newblock ``Image quality assessment: Unifying structure and texture
  similarity,''
\newblock {\em IEEE transactions on pattern analysis and machine intelligence},
  2020.

\bibitem{talebi2018nima}
Hossein Talebi and Peyman Milanfar,
\newblock ``Nima: Neural image assessment,''
\newblock {\em IEEE transactions on image processing}, 2018.

\bibitem{wang2021rethinking}
Pei Wang, Yijun Li, and Nuno Vasconcelos,
\newblock ``Rethinking and improving the robustness of image style transfer,''
\newblock in {\em Proceedings of the IEEE/CVF Conference on Computer Vision and
  Pattern Recognition}, 2021.

\bibitem{gal2022stylegan}
Rinon Gal, Or~Patashnik, Haggai Maron, Amit~H Bermano, Gal Chechik, and Daniel
  Cohen-Or,
\newblock ``Stylegan-nada: Clip-guided domain adaptation of image generators,''
\newblock {\em ACM Transactions on Graphics (TOG)}, 2022.

\bibitem{patashnik2021styleclip}
Or~Patashnik, Zongze Wu, Eli Shechtman, Daniel Cohen-Or, and Dani Lischinski,
\newblock ``Styleclip: Text-driven manipulation of stylegan imagery,''
\newblock in {\em Proceedings of the IEEE/CVF International Conference on
  Computer Vision}, 2021.

\end{thebibliography}
\end{document}